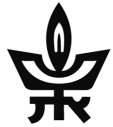

# TEL AVIV UNIVERSITY

# The Raymond and Beverly Sackler faculty of Exact Sciences

# School of Mathematical Sciences

# Exclusive Row Biclustering for Gene Expression Using a Combinatorial Auction Approach

A thesis submitted toward the degree of

Master of Science in Statistics

by

# Amichai Painsky

This research was carried out
in the Department of Statistics and Operations Research under
the supervision of Professor Saharon Rosset

March   2012

# Abstract

The availability of large microarray data has led to a growing interest in biclustering methods in the past decade. Several algorithms have been proposed to identify subsets of genes and conditions according to different similarity measures and under varying constraints. In this paper we focus on the exclusive row biclustering problem for gene expression data sets, in which each row can only be a member of a single bicluster while columns can participate in multiple ones. This type of biclustering may be adequate, for example, for clustering groups of cancer patients where each patient (row) is expected to be carrying only a single type of cancer, while each cancer type is associated with multiple (and possibly overlapping) genes (columns). We present a novel method to identify these exclusive row biclusters through a combination of existing biclustering algorithms and combinatorial auction techniques. We devise an approach for tuning the threshold for our algorithm based on comparison to a null model in the spirit of the Gap statistic approach [20]. We demonstrate our approach on both synthetic and real-world gene expression data and show its power in identifying large span non-overlapping rows sub matrices, while considering their unique nature. The Gap statistic approach succeeds in identifying appropriate thresholds in all our examples.

# Table of Contents

# 1   Introduction

The technology of DNA chips (microarrays) provides an effective tool of gene expression mRNA level measurement in different conditions. This technology allows a comprehensive overview of genes' transcriptional behavior under different environmental conditions and provides a powerful source of information for our understanding of biological systems. Gene expression data is typically arranged in a matrix form, where each entry of the matrix is usually the logarithm of the relative abundance of the gene's mRNA under a specific condition. From the gene expression matrix we would like to extract valuable information on the way subsets of genes behave under sets of conditions. We would therefore like to identify submatrices (biclusters) of the gene expression matrix, in a way that each bicluster represents a group of genes that behave in a similar manner (pattern) under a set of conditions. The biclustering problem was tackled by many researchers who suggested a broad range of methods and algorithms. These methods span different bicluster types, bicluster structures, different quality measures and different heuristics to overcome the computational complexity of the biclustering problem. A comprehensive survey which encompasses a vast majority of the work that has been done was presented by Madeira and Oliveira [1].

In this work we focus on a specific type of biclustering problems, in which we aim to identify maximal volume biclusters such that a row can only be member of a single bicluster (exclusive row). This type of biclustering is especially interesting in gene expression problems where each individual (represented by a row) is assumed to be a member of only a single bicluster, while different biclusters may be associated with the same genes (columns). As an example, given a group of patients where each patient is known to be carrying only a single type of Leukemia, we would like to cluster the patients according to their Leukemia type and at the same time discover those genes that demonstrate a unique pattern for each type of Leukemia. Through accurate exclusive row biclustering we enhance the study of genes' transcriptional behavior in cases where we are able to collect genetic fingerprints from different individuals, and each individual is known to be influenced by a single, well-differentiating, genetic related phenomenon.

We start with a short review of microarray biclustering types and structures, focusing on our specific problem setup. We present our formal problem statement and discuss our suggested solution approach. We then introduce our algorithm for exclusive-row biclustreing given a threshold on the MSR, as well as our Gap statistic based approach for tuning the threshold. We examine the performance of our method on both synthetic and real-world Leukemia gene expression data. We discuss our experiments' results and compare our approach's performance with previously proposed methods. We show that our approach demonstrates significantly better results in both types of experiments, compared with an overlap tolerant MSR based method [3] in the synthetic data experiments, and with an exclusive row method, based on a Gibbs sampling approach [5], in the Leukemia gene expression experiment. In particular, the MSR threshold selected by the Gap approach leads to favorable performance in all examples.

## 1.1 Bicluster Type

Different types of biclusters can be assumed to exist in a gene expression microarray. Among the common types are:

1. Biclusters with constant values.

2. Biclusters with constant values on rows or columns.

3. Biclusters with coherent values.

4. Biclusters with coherent evolutions.

| 2 | 2 | 2 | 2 | 2 |
|---|---|---|---|---|
| 2 | 2 | 2 | 2 | 2 |
| 2 | 2 | 2 | 2 | 2 |
| 2 | 2 | 2 | 2 | 2 |
| 2 | 2 | 2 | 2 | 2 |

(a)

| 2 | 2 | 2 | 2 | 2 |
|---|---|---|---|---|
| 3 | 3 | 3 | 3 | 3 |
| 1 | 1 | 1 | 1 | 1 |
| 4 | 4 | 4 | 4 | 4 |
| 5 | 5 | 5 | 5 | 5 |

(b)

| 1 | 3 | 2 | 4 | 5 |
|---|---|---|---|---|
| 1 | 3 | 2 | 4 | 5 |
| 1 | 3 | 2 | 4 | 5 |
| 1 | 3 | 2 | 4 | 5 |
| 1 | 3 | 2 | 4 | 5 |

(c)

| 1   | 3   | 5   | 7    | 9    |
|-----|-----|-----|------|------|
| 1.5 | 3.5 | 5.5 | 7.5  | 9.5  |
| 3.5 | 5.5 | 7.5 | 9.5  | 11.5 |
| 4.5 | 6.5 | 8.5 | 10.5 | 12.5 |
| 2   | 4   | 6   | 8    | 10   |

(d)

| $a_1$ | $a_1$ | $a_1$ | $a_1$ | $a_1$ |
|-------|-------|-------|-------|-------|
| $a_2$ | $a_2$ | $a_2$ | $a_2$ | $a_2$ |
| $a_3$ | $a_3$ | $a_3$ | $a_3$ | $a_3$ |
| $a_4$ | $a_4$ | $a_4$ | $a_4$ | $a_4$ |
| $a_5$ | $a_5$ | $a_5$ | $a_5$ | $a_5$ |

(e)

| $a_1$ | $a_2$ | $a_3$ | $a_4$ | $a_5$ |
|-------|-------|-------|-------|-------|
| $a_1$ | $a_2$ | $a_3$ | $a_4$ | $a_5$ |
| $a_1$ | $a_2$ | $a_3$ | $a_4$ | $a_5$ |
| $a_1$ | $a_2$ | $a_3$ | $a_4$ | $a_5$ |
| $a_1$ | $a_2$ | $a_3$ | $a_4$ | $a_5$ |

(f)

**Figure 1.** *Bicluster types: (a) constant values, (b) constant values on rows, (c) constant values on columns, (d) coherent values, (e) coherent evolutions on rows, and (f) coherent evolutions on columns*

The first three bicluster types directly analyze the numeric values in the data matrix and try to find subsets of rows and subsets of columns with similar behaviors. These behaviors can be observed on the rows, on the columns, or in both dimensions of the data matrix, as in figure 1.a, 1.b, 1.c and 1.d The fourth class aims to find coherent behaviors regardless of the exact numeric values in the data matrix. As such, biclusters with coherent evolutions view the elements in the data matrix as symbols.

In the case of gene expression data, constant biclusters (figure 1.a) reveal subsets of genes with similar expression values within a subset of conditions. A bicluster with constant values in the rows (figure 1.b) identifies a subset of genes with similar expression values across a subset of conditions, allowing the expression levels to differ from gene to gene. Similarly, a bicluster with constant columns (figure 1.c) identifies a subset of conditions within which a subset of genes present similar expression values assuming that the expression values may differ from condition to condition. However, one can be interested in identifying more complex relations between the genes and the conditions by looking directly at the numeric values or regardless of them. As such, a bicluster with coherent values (figure 1.d) identifies a subset of genes and a subset of conditions with coherent values on both rows and columns. This bicluster type will be discussed in more detail in the next section. On the other hand, identifying a bicluster with coherent evolutions (figure 1.e) may be helpful if one is interested in finding a subset of genes that are up-regulated or down-regulated across a subset of conditions without taking into account their actual expression values; or if one is interested in identifying a subset of conditions that have always the same or opposite effects on a subset of genes.

According to the specific properties of each problem, one or more of these different types of biclusters is generally considered interesting. Moreover, a different type of merit function should be used to evaluate the quality of the biclusters identified. The choice of the merit function is strongly related with the characteristics of the biclusters each algorithm aims to find.

## 1.2  Mean Square Residue

Cheng and Church [2] were the first to introduce the concept of biclustering. They defined the MSR as their biclusters' quality measure. We follow their notation through the remaining sections of our work. The gene expression matrix is noted as $A = (X, Y)$ with a set of rows X and a set of columns Y. Each entry of the matrix is noted as $a_{ij}$ and corresponds to the $i^{th}$ row and $j^{th}$ column. We represent a bicluster as a submatrix $(I, J)$ where $I \subset X$ and $J \subset Y$. We define $a_{iJ}$ to be the mean of $i^{th}$ row over all columns $J$ of the bicluster. In a similar manner $a_{Ij}$ is the mean of the $j^{th}$ column over all rows $I$ in the bicluster. $a_{IJ}$ is the mean of all entries of the bicluster $(I, J)$.

$$a_{iJ} = \frac{1}{|J|} \sum_{j \in J} a_{ij}. \qquad (1)$$

$$a_{Ij} = \frac{1}{|I|} \sum_{i \in I} a_{ij}. \qquad (2)$$

$$a_{IJ} = \frac{1}{|I||J|} \sum_{j \in J, i \in I} a_{ij} \, . \tag{3}$$

For our problem setup we are interested in coherent values biclusters. According to an additive model, a perfect bicluster (I, J) with coherent values is defined as a subset of rows and columns such that each entry of the submatrix can be expressed as in equation (4).

$$\bar{a}_{ij} = \mu + \alpha_i + \beta_j \, . \tag{4}$$

Where $\mu$ is a typical constant value of the bicluster, $\alpha_i$ is an offset for the row $i \in I$, and $\beta_j$ is an offset for the column $j \in J$. figure 1.d shows an example of an additive model coherent bicluster.

In order to quantify how perfect a bicluster is, Cheng and Church [2] also defined a residue element $r_{ij}$ which corresponds to the difference between the real measured value of an entry in the bicluster $a_{ij}$ and its *perfect* value according to the additive model (5).

$$r(a_{ij}) = a_{ij} - \bar{a}_{ij} = a_{ij} - a_{iJ} - a_{Ij} + a_{IJ} \, . \tag{5}$$

To quantify the quality of the entire bicluster we define the MSR (6).

$$H(I, J) = \frac{1}{|I||J|} \sum_{i \in I, j \in J} r(a_{ij})^2 \, . \tag{6}$$

We say that a bicluster (I, J) is a δ-bicluster if $H(I, J) < \delta$.

The MSR is commonly used when quantifying the quality of coherent value biclusters [1-4]. We shall therefore use it in our paper as our quality measure, as described in the following sections.

### 1.3 Bicluster Structure Review

Biclustering algorithms are designed to extract either a single or K multiple biclusters from a microarray dataset, where K (parameter) is the number of biclusters we expect to identify and is usually defined a-priori.

When the biclustering algorithm assumes the existence of several biclusters in the data matrix, the following bicluster structures can be obtained:

1. Exclusive rows and columns biclusters.

2. Non-overlapping biclusters with checkerboard structure.

3. Exclusive rows biclusters.

4. Exclusive columns biclusters.

5. Overlapping biclusters with hierarchical structure.

6. Arbitrarily positioned overlapping biclusters.

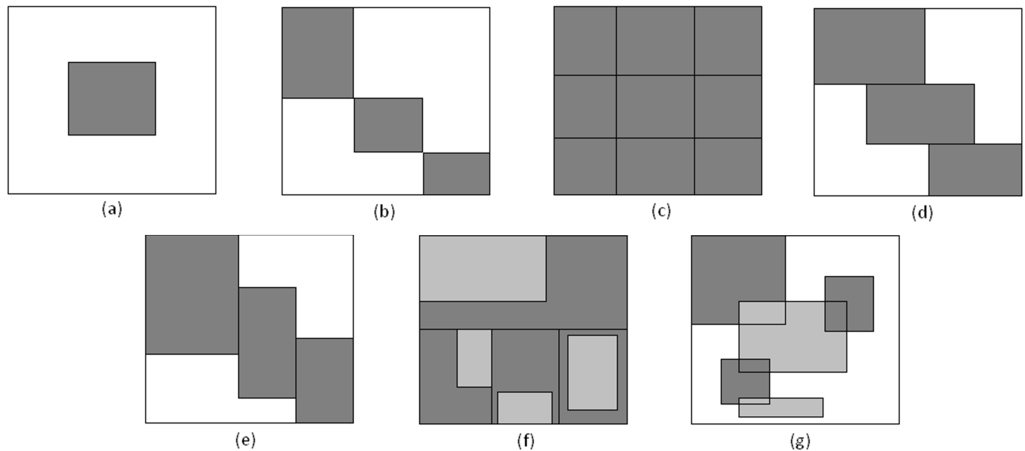

**Figure 2.** *Bicluster structures: (a) single bicluster, (b) exclusive rows and columns biclusters, (c) non-overlapping biclusters with checkerboard structure, (d) exclusive rows biclusters, (e) exclusive columns biclusters, (f) overlapping biclusters with hierarchical structure, and (g) arbitrary positioned overlapping biclusters*

Figure 2.a demonstrates a single bicluster, located at the middle of the matrix and represented by a gray shaded area.

Figure 2.b shows an example of K exclusive row and column biclusters (where K=3), with every row and every column in the matrix belongs exclusively to one of the K biclusters.

An example of a non-overlapping biclusters with checkerboard structure is presented in figure 2.c, where this unique structure considers that rows and columns may belong to more than one bicluster, and assume a checkerboard structure in the data matrix. By doing this, we allow the existence of K non-overlapping and nonexclusive biclusters where each row in the data matrix belongs to exactly K biclusters. The same applies to columns.

Other approaches assume that rows can only belong to one bicluster, while columns, which correspond to conditions in the case of gene expression data, can belong to several biclusters. This structure is presented in figure 2.d

This approach can also produce exclusive-columns biclusters when the applied on the opposite orientation of the data matrix. When this is the case, the columns can only belong to one bicluster while the rows can belong to one or more biclusters. This structure is then called exclusive-columns biclusters (see figure 2.e).

The structures presented in figure. 2.b, 2.c, 2.d, and 2.e assume that the biclusters do not overlap. However it is more likely that, in real data, some rows or columns do not belong to any bicluster at all and that the biclusters overlap in some places.

Figure 2.f shows an example of overlapping biclusters with hierarchical structure, where every bicluster is a subsetsof a single bicluster and doesn't overlap with other biclusters besides its ancestor.

All the bicluster structures discussed above share an exhaustive biclusters structure property. That is, that every row and every column belongs to at least one bicluster. However, we can consider non-exhaustive variations of these structures that make it possible that some rows and columns do not belong to any bicluster, which may apply to more sophisticated (and realistic) setups.

Figure 2.g demonstrates a more general bicluster structure allows the existence of K possibly overlapping biclusters without taking into account their direct observation on the data matrix with a common reordering of its rows and columns. Furthermore, these nonexclusive biclusters can also be non-exhaustive, which means that some rows or columns may not belong to any bicluster.

## 1.4 Exclusive row biclusters

As discussed above several structures of biclusters can be assumed to exist in a given gene expression matrix, as they are well described in Madeira and Oliveira's comprehensive survey [1]. In this paper we focus on the simultaneous extraction of multiple exclusive row biclusters. Exclusive row biclusters are biclusters which may overlap in their columns but not in their rows. As most biclustering methods focus on overlap or non-overlap grouping in both dimensions, we suggest that in some applications we have an interest in exclusively assigning a row element to only a single group, while columns can be viewed as explanatory variables and simultaneously play role in multiple biclusters. Our goal is therefore to find maximal size biclusters, such that as many row elements as possible will be grouped in a non-overlap manner, while being explained by as many columns as possible. Although limited previous work exists that directly formulates and solves the exclusive-row biclustering problem, several previous methods can be related to this problem. Sheng et al. [5] introduced a Gibbs Sampling based exclusive row biclustering method. Their approach considers sequential extraction of coherent value biclusters, where clustered rows are masked for future iterations, to prevent overlap. Tang et al. [6] suggested an Interrelated Two Way Clustering (ITWC) approach, which focuses on the extraction of constant value ($\alpha_i = \beta_j = 0$) biclusters by iteratively applying a one dimensional clustering algorithm on each of the dimensions. Divina et al. presented the SEBI Evolutionary Computation based algorithm which aims to find

MSR based biclusters with a low level of overlap. They suggested simultaneous discovery of biclusters and added the low level overlap condition in their element's weight $w_p(a_{ij})$ as described in detail in [7]. In this paper, we present a novel method for simultaneous discovery of exclusive row coherent value biclusters, based on a simple relaxation of an optimization problem presented in the next section. We use Sheng et al. work as a comparison reference for our method, as it also aims to discover coherent value exclusive row biclusters, based on a different approach than we do.

## 2 Methods

### 2.1 Problem Formulation

As stated above, we are interested in finding maximal size δ-biclusters, such that each row element can only be assigned to a single biclusters (no overlap in rows), while no such constraint holds for the columns. We would also like to define a minimal number of rows and columns in each biclusters, to prevent the solution from containing single element biclusters. Formally, for a matrix $A = (X, Y)$, an MSR threshold δ and minimal bicluster dimensions m and n, we would like to find a set of biclusters $(I, J)_k$ such that:

$$(I, J)_k = \mathrm{argmax}_{I_k \subset X, J_k \subset Y} \sum_k |I_k| \cdot |J_k| \qquad (7)$$

$$\text{subject to:} \quad H(I_k, J_k) < \delta \qquad \forall k$$

$$|I_k| > m, |J_k| > n \qquad \forall k$$

$$I_i \cap I_j = \emptyset \qquad \forall i \neq j$$

It is easy to show that this optimization problem is NP-hard and an optimal solution cannot be achieved in a non-exhaustive search method, which may take an exponential time to compute. We therefore suggest a relaxation to the problem, by solving it in two stages:

- Stage 1 – Optimization without the non-overlapping biclusters constraint:

$$(\tilde{I}, \tilde{J})_k = \mathrm{argmax}_{I_k \subset X, J_k \subset Y} \sum_k |I_k| \cdot |J_k| \qquad (8)$$

$$\text{subject to:} \quad H(I_k, J_k) < \delta \qquad \forall k$$

$$|I_k| > m, |J_k| > n \qquad \forall k$$

$$(I_i, J_i) \neq (I_j, J_j) \qquad \forall i \neq j$$

- Stage 2 – Adding the no-overlap constraint:

$$(I, J)_k = \text{argmax}_{I_k \subset \bar{I}, J_k \subset \bar{J}} \sum_k |I_k| \cdot |J_k| \tag{9}$$

$$\text{subject to:} \quad I_i \cap I_j = \emptyset \quad \forall i \neq j$$

Notice that in the first stage our goal is to find δ-biclusters such that their size is as large as possible and with no limitation on the number of biclusters we are looking for. In other words, in the first stage our algorithm attempts to find all biclusters that hold the specified constraints and may overlap, but are not exactly the same. The tendency is to find the largest possible ones, according to the optimization objective. In the second stage our algorithm searches over the biclusters it found in its previous stage, and chooses the ones that achieve maximal volume without row overlap. Notice that each stage of the relaxed problem is still an NP-hard optimization problem. However these problems can be formulated so that they can be solved by well-studied approximations, and the combination of the two stages achieves a solution to the relaxed problem, which is only dependent on the quality of each of the approximations.

## 2.2 Stage 1: Mean Square Residue Based Biclustering

For the purposes to this work we focus on the Flexible Overlapped Biclustering (FLOC) algorithm, introduced by Yang et al. [3], as a biclustering method for our first stage. In their work, Yang et at. presented a method for simultaneous discovery of K δ-biclusters, such that their volume is maximized. They defined *an action* with respect to a bicluster as an addition/removal of a row/column to a bicluster. The *gain* of an action, denoted Gain(X,C), was therefore defined as a combination of residue reduction and an increase of volume, caused by an action X on bicluster C:

$$\text{Gain}(x, c) = \frac{r - r_{c'}}{\frac{r^2}{r_c}} + \frac{v_{c'} - v_c}{v_c} \tag{10}$$

Where x is a row/column addition/removal operation, c is a bicluster, c′ is the bicluster obtained by performing operation x on c, r is the MSR threshold (which we previously defined δ), $r_c$ and $r_{c'}$ are the residues of biclusters c and c′ respectively and $v_c, v_{c'}$ are their volumes. Notice that the gain obtained by an action on a bicluster is influenced by both its residue and volume, and increases as a result of a residue decrease, or a volume increase. Further analysis of $\text{Gain}(x, c)$ is discussed in [3].

The FLOC algorithm operates in two phases, as presented in figure 3. In the first phase it initializes K biclusters by randomly assigning elements to each of them. It then continues to the second phase in which it chooses for each row and column the best *action* with respect to each of the K biclusters. In other words, at each iteration, the algorithm will go over all rows and columns, and for each of them it will look for the highest Gain(X,C) where X is addition/removal operation of that row/column and C is one of the K bicluster. The algorithm applies the operation which results with the maximal Gain(X,C) for each of the rows and columns. Several heuristics are applied on the calculation of the gain of an action to reduce its computational complexity, as described in [3].

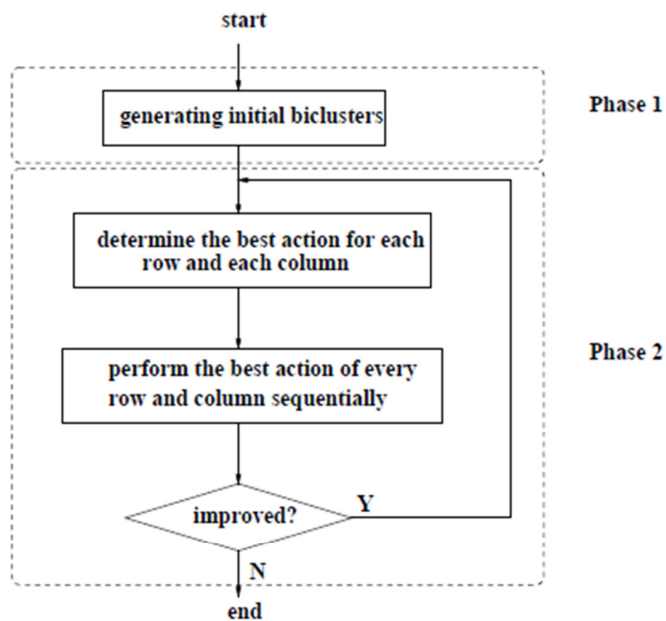

**Figure 3.** *FLOC flow chart.*

Yang et al. also suggested several improvements to their original algorithm [4] which include initialization of different biclusters according to different random draws and dynamic order of actions (in its original version, the order the algorithm goes over the rows and columns at each iteration is predetermined and has a major effect on the resulting biclusters). Refer to [4] for a comprehensive review.

Notice that The FLOC algorithm has no overlap avoidance component and may find many extremely overlapped δ-biclusters, depending on their initialization and the given MSR threshold.

## 2.3 Stage 2: Adding the Non-overlap Constraint on Rows

We formulate our stage 2 problem as a combinatorial auction problem.

### 2.3.1 Combinatorial Auctions

A combinatorial auction is a type of an auction in which bidders (participants) can place bids on bundles of goods rather than individual items (as in traditional auctions). The growing interest in the past years in such auction mechanisms stemmed from resource allocation problems in competitive multi-agent systems. In a combinatorial auction, each bidder places bids on bundles of goods, according to how valuable these bundle are for him. The goods are then auctioned simultaneously and bidders place as many bids as they want for different bundles of goods with the guarantee that these bundles will be allocated in an "all-or-nothing" manner. This means either the bidder wins the entire bundle, or he doesn't win the bundle at all. A bidder cannot win a partial bundle. Particularly, we are interested in the winner determination problem (WDP). In a combinatorial auction, a seller is faced with a set of price offers for various bundles of goods. His goal is to allocate the goods in a way that maximizes his revenue. The WDP is defined as choosing the subset of bids that maximizes the seller's revenue, subject to the constraint that each good can be allocated at most once. Formally, the WDP is the following integer program:

$$\max \sum x_i p(b_i) \qquad (11)$$

$$\text{subject to: } \sum_{i | \gamma \in g(bi)} x_i \leq 1 \quad \forall \gamma \in G$$

$$x_i \in \{0,1\} \qquad \forall i$$

Where $G = \{\gamma_1, \gamma_2, \ldots, \gamma_m\}$ is the set of goods, $B = \{b_1, b_2, \ldots, b_n\}$ is the set of bids, a bid $b_i$ corresponds to the pair $(p(b_i), g(b_i))$ where $p(b_i)$ is a non negative price offer, and $g(b_i)$ is the set (bundle) of goods requested by $b_i$. We also define $x_i$ to be an indicator function which takes the value 1 or 0 if the bid $b_i$ was won (or not). Rothkopf et al. show that the WDP is equivalent to a weighted set-packing problem and is therefore NP-hard [9]. Furthermore, it can be shown that the WDP is inapproximable within any constant factor [10]. In recent years many researchers have been interested in the WDP, as reviewed in de Vries et al. survey [11]. A major focus was given on finding tractable subcases [9, 11-13] and approximation algorithms, despite the lack of approximation guarantees [12, 14].

Although the WDP is asymptotically NP-hard, in practice it is possible to address interestingly-large datasets with heuristic methods [10, 15-17], in order to find an optimal solution for the WDP.

For our stage 2 optimization problem, we look at the δ-biclusters we found in stage 1 as bidders in a WDP, bidding for bundles of goods (rows), where the bidding price is the volume of the δ-bicluster. We explain and demonstrate this process in detail in Section 2.4 below.

### 2.3.2 The CASS Algorithm

The algorithm we choose to focus on for the purposes of this work is the branch and bound based Combinatorial Auction Structured Search (CASS) algorithm [8]. CASS is an exhaustive search algorithm which considers fewer partial allocations than the brute-force methods. It structures the search space in a way that provides context to this heuristic in order to allow more pruning during the search and that avoids consideration of most infeasible allocations. CASS also caches the results of partial searches and prunes the search tree. CASS is based on a Branch and Bound search mechanism. Whenever a bid is encountered and does not conflict with the current partial allocation, the search tree branches. Then one branch adds the bid to the partial allocation while the other does not. CASS performs a depth-first search, meaning that one branch of the tree is fully explored before the other is considered. This has the advantage that CASS requires only linear space to store the search tree. When a full allocation is reached CASS records this allocation then backtracks. CASS also computes a revenue upper bound function at each node of the search tree. This upper bound refers to the revenue that can be collected from the goods that are not part of the current branch's allocation. It is used to indicate if this branch of the tree may lead to a solution better than the one we currently hold. Other mechanisms, such as bins, caching and bid ordering heuristics are used to enhance the performance of the CASS algorithm and described in detail in [8]. The CASS algorithm is guaranteed to converge to the optimal WDP allocation. Experiments show that for reasonably large WDPs and sufficient computational resources, which are adequate for our problem, it provides the optimal allocation quite rapidly [8].

## 2.4 Exclusive Row Mean Square Residue Biclustering

Going back to our initial objective, given a matrix $A = (X, Y)$, we would like to find maximal size δ-biclusters which do not overlap in their rows. As described above, we relax our optimization problem by solving it in two separate stages.

- Stage 1 – Optimization without the non-overlapping biclusters constraint

In the first stage our algorithm attempts to extract many large volume δ-biclusters without the overlap constraint. We would like to discover as many δ-biclusters as possible, as they will be used as candidates for exclusive row optimization in the next stage. To do so, we apply the FLOC algorithm on our input matrix $A = (X, Y)$. We use an R implementation of the algorithm, provided by the BicARE package [19]. Every time it is applied, the FLOC algorithm discovers a configurable number (K) of δ-biclusters. These biclusters are not guaranteed to be non-exclusive (and in some cases may be very similar), nor to converge to the optimal solution as FLOC performs a heuristic greedy search. Therefore, we apply the FLOC algorithm multiple times, at different random initializations, in order to generate as large a variety of different δ-biclusters as possible. Moreover, as the algorithm tends to discover large volume δ-biclusters, we apply it with δ values smaller than the one we are requested to. By doing so, we also generate smaller size biclusters which naturally still hold the original MSR constraint. The role of these smaller volume δ-biclusters is discussed in the next paragraph.

- Stage 2 - Exclusive row volume maximization

Given a set of overlapping δ-biclusters we are ready to choose the optimal subset such that their global volume is maximal and without row overlap. This problem can be formulated as a combinatorial auction WDP and solved by applying the CASS algorithm on it.

As mentioned above, we consider the δ-biclusters we found in the previous section as bidders in a WDP, bidding for bundles of goods (rows), where the bidding price is the volume of the δ-bicluster. This means that every δ-bicluster bids on the rows that construct it, and its bidding price is set to be its volume. The WDP attempts to maximize the combinatorial auction's revenue, which is the sum of submitted prices (volumes) in a way that a single good (row) cannot be allocated to multiple bidders (δ-biclusters). Since we perform an "all-or-nothing" combinatorial auction, in which a bidder can either win its entire bundle or does not win it at all (no partial bundles allocation allowed) we would like to also have smaller size biclusters. These smaller biclusters, which naturally bid on a smaller number of rows, may win them in case bigger volume biclusters fail to do so. This way we overcome the problem of non-clustered rows, caused by conflicting large bundles, as demonstrated in figure 4.

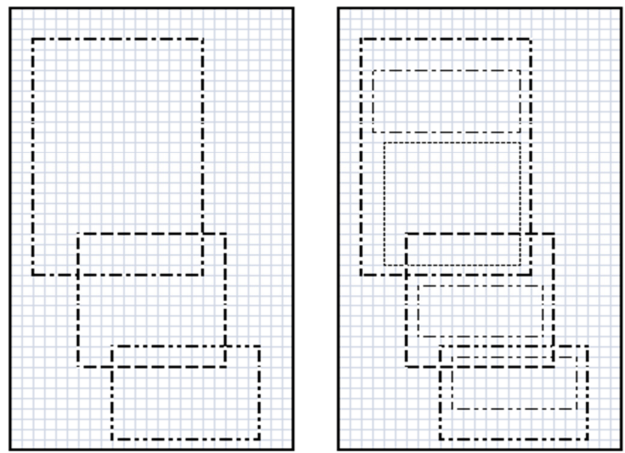

**Figure 4.** *Left: biclustering with a given MSR threshold. Notice that the CASS algorithm would discard the middle bicluster in the second stage of our method. This results with quite a large portion of non-clustered rows. Right: here we illustrate the advantage of searching for biclusters with smaller values of MSR threshold (represented with a thinner dashed border) in addition to the original MSR threshold constraint. By doing so we generate more candidate biclusters which still hold all constraints and maximize the global volume discovered by our complete algorithm, after CASS is applied.*

Our complete algorithm can now be presented:

For a given matrix A=(X,Y), an MSR threshold δ>0, and minimal bicluster dimensions m>0, n>0:

Stage 1

a. Apply the FLOC biclustering algorithm on the Matrix A, with the given parameters δ,m,n. Choose K (number of searched the δ-biclusters) to be as computationally large as possible, and randomly initialize the K biclusters.
b. Repeat step (1) with smaller values of δ, to also achieve smaller biclusters which still comply with all constraints.

Stage 2

a. Calculate the volume of each of the biclusters found in the previous stage.
b. Apply the CASS algorithm on the biclusters such that:
   i. Biclusters are bidders.
   ii. Rows are goods.
   iii. Bundles' biding prices are the biclusters volume.
c. The winners of the combinatorial auction WDP will be the set of exclusive row δ-biclusters which maximize the sum of volumes.

## 2.5 Complexity Analysis

Our method consists of two stages, with cascaded and separate algorithms. The first stage applies the FLOC biclustering algorithm on a given $A = (X, Y)$ matrix. It is shown in [3] that its computational complexity is bounded by $O((|X| + |Y|)^2 \cdot K \cdot p)$ where $|X|, |Y|$ are the dimensions of matrix A, K is the number of searched δ-biclusters and p is the maximal number of iterations before termination.

The CASS algorithm, on the other hand, demonstrates an exponentially increasing complexity with the number of goods (rows, $|X|$) [8]. However, due to its heuristics and implementation optimizations, it proves to converge quite rapidly for reasonably large dataset, such as in our problem, as described in [8]; our experiments show CASS takes less than tenth of our overall runtime.

## 2.6 MSR threshold selection

A major challenge of any clustering approach is the selection of an "optimal" (or at least good) threshold for "homogeneity" of the resulting clusters, to give results that are most meaningful. In a standard one-way clustering, a widely used approach is Tibsirani et al. Gap statistic [20]. We now introduce this approach and adapt it to our setup of exclusive-row biclustering using the MSR as a homogeneity measure.

### 2.6.1 Estimating the number of clusters via Gap Statistic

In their paper, Tibshirani et al. discuss the challenge of estimating the optimal number of clusters K when applying almost any clustering method on a given dataset. They show that in standard clustering, the error (non-homogeneity) measure tends to monotonically decrease as the number of clusters increase, but from some K on the decrease flattens markedly. This K is usually referred to as the elbow of the plot and is believed to indicate the optimal number of clusters K in the data set.

The Gap Statistics method provides a statistical procedure to formularize the detection of the elbow. The idea of this approach is to standardize the graph of $\log(W_k)$ (the error measure, defined below) by comparing it to its expectation under an appropriate null reference of the data set.

In their paper, Tibshirani et al. define the within cluster dispersion $W_k$ as:

$$W_k = \sum_{r=1}^{k} \frac{1}{2n_r} D_r \qquad (12)$$

Where n is the number of rows (observations), k is the number of clusters and $D_r$ is the sum of pairwise distances in cluster r:

$$D_k = \sum_{i,i' \in C_r} d_{ii'} \qquad (13)$$

Plotting the logarithm of $W_k$ against the number of clusters K shows a monotonically decreasing behavior, with an elbow that is believed to represent the optimal number of clusters in the data set, as demonstrated in figure 5.

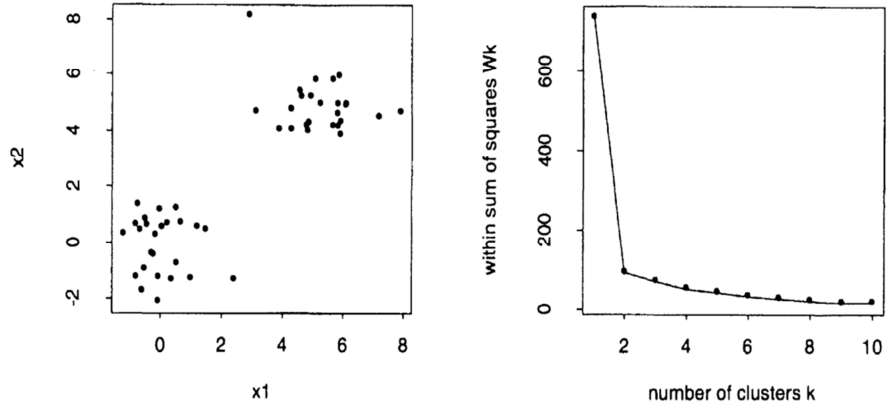

**Figure 5.** *Two clusters example. Left: two dimensional data samples. Right: Monotonically decreasing behavior of $W_k$ as function of k , with an elbow at k=2 suggesting the optimal number of clusters in the data.*

Tibshirani et al. defined the Gap statistic as:

$$\text{Gap}_n(k) = E_n^* \log(W_k) - \log(W_k). \tag{14}$$

Where $E_n^*$ denotes the expectation under a sample of size n from a null reference distribution with only one cluster in it. The estimate $\hat{k}$ is the value maximizing $\text{Gap}_n(k)$.

In their framework, Tibshirani et al. adopt a null model of a single component, and compare it to the models derived by the clustering for all values of k. They seek the value of k which supplies the "strongest" evidence against the null and select it as the number of clusters.

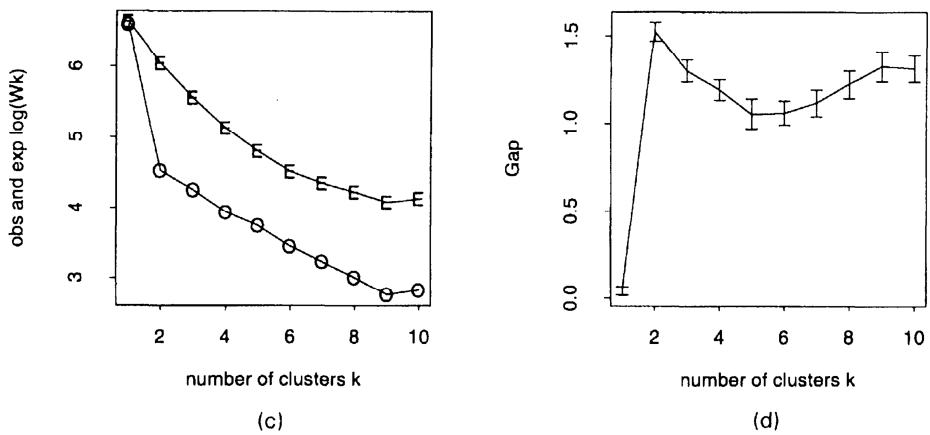

(c)            (d)

**Figure 6.** *Two clusters example for Tibshirani et al. Gap statistic (continuing figure 5). Left: $\log(W_k)$ (O) and $E_n^* \log(W_k)$ (E) as function of k. Right: $\text{Gap}_n(k)$ with $\hat{k} = 2$ as expected.*

Tibshirani et al. justified the use of the logarithm of $W_k$ by showing that when clustering n uniformly distributed data points with K centers, and assuming that the centers are aligned in an equally spaced fashion, then the Gap statistic achieves its maxima exactly at $\hat{k} = K$. Additionally, they suggested that in the case of a special Gaussian mixture model, $\log(W_k)$ has an interpretation as a log-likelihood.

On the other hand, Mohajer et al. [21] show that using the logarithm of $W_k$ (instead of $W_k$ directly) may be disadvantageous in several other cases, especially in cases of multidimensional uniform distribution with large difference in variances of the different clusters. We concur with Mohajer et al. in concluding that the decision on whether or not to use the logarithm function shall be tailored to the specific problem setup, and shall be based eventually on empirical means, due to the complexity of the problem.

### 2.6.2 Estimating the MSR threshold via Gap Statistic

We follow the footsteps of the Tibshirani et al.'s approach to introduce an MSR threshold estimation method, based on the Gap statistic idea.

Looking at our optimization problem (7), we notice that increasing the MSR threshold value results with greater volume discovered; step 1.b. of our suggested method, together with the optimality of the CASS algorithm, guarantees that the volume discovered for any MSR threshold value will necessarily be greater or equal to the volume discovered when applying smaller threshold values. In other words, the total volume discovered by our suggested method is a monotonically non-decreasing function of the MSR threshold value.

Assume a null hypothesis data set is drawn from a certain distribution, noted as the reference distribution, and construct a data set by embedding several biclusters with a known MSR in it. As we start to increase $\delta_{th}$ we discover more of the embedded biclusters. However, as we increase $\delta_{th}$ above the embedded biclusters' MSR value, we achieve larger biclusters than the ones we embedded, due to the addition of data entities that were drawn from the reference distribution and are not part of the embedded biclusters. In other words, increasing the MSR threshold above the embedded biclusters' MSR value shall result with a volume increase rate that is typical to the dataset's reference distribution. In order to detect this MSR threshold value, which is indicated by the elbow of the graph, we apply a Gap statistic method, comparing the discovered volume detected in the dataset with the volume discovered in reference dataset, drawn from the reference distribution.

We define the Gap statistic as:

$$\text{Gap}_n(\delta_{th}) = V(\delta_{th}) - E_n^* V(\delta_{th}). \tag{15}$$

Where $V(\delta_{th})$ is the volume discovered for an MSR threshold value $\delta_{th}$ and $E_n^*$ denotes expectation under a sample of size n from the reference distribution. This formulation of the Gap statistic, without the logarithm, is discussed in detail in [21]. Our experiments show that the direct use of the volume to describe the Gap statistic is suitable for both synthetic and real world microarray datasets.

### 2.6.3 Gap statistic implementation

In order to implement the Gap statistic we first need to define a method for choosing a reference distribution.

Looking at the definition of $V(\delta_{th})$, we see that we achieve greater volume by loosening the MSR constraint and vice versa. In other words, looking for a reference distribution that will maximize (minimize) the expectation of the discovered volume for a given MSR constraint is equivalent to finding such distribution that will minimize (maximize) the expectation of the MSR, given the bicluster's dimensions. Deriving the expectation of the MSR for a given A(I,J) it is easy to show (Theorem 1) that it is only depends on its second moments, E $a_{ij}a_{mn}$     $\forall i, m \in I, j, n \in J$.

Therefore, $E_n^* V(\delta_{th})$ also only depends on the chosen reference distribution through its second order moments, and not through its entire distribution function. However, we notice that by choosing a reference distribution with zero variance we achieve maximal $E_n^* V(\delta_{th})$ for any threshold value (the entire volume of the matrix), and a distribution with an extremely large variance that will results with $E_n^* V(\delta_{th}) \approx 0$.

In other words, the question of how we choose a reference distribution for $E_n^* V(\delta_{th})$ is reduced to finding an appropriate variance constraint for a chosen reference distribution.

We use the assumption that for each column, entries that are not members of any embedded biclusters were independently drawn from the same distribution (a reasonable assumption for our microarray dataset, where rows are patients and columns are genes). Therefore, our variance estimation problem reduces to one of estimating the variance of each column according to its non-clustered entries. The resulting algorithm for estimating the Gap statistic is therefore:

For a given MSR threshold, $\delta_{th}$>0:

1. Compute $\mathbf{V(\delta_{th})}$, the total volume discovered by applying our exclusive row biclustering algorithm applied on the data matrix A(I,J).

2. Compute the variance of each column from its observed values, excluding those entries that are biclusters' members, according to the biclusters discovered in (1).

3. Construct a reference dataset, drawn from a uniform distribution according to the variance constraint computed in (2).

4. Apply the exclusive row biclustering algorithm on the reference dataset.

5. Repeat steps (3) and (4) to achieve $\mathbf{E_n^* V(\delta_{th})}$.

6. Subtract $\mathbf{E_n^* V(\delta_{th})}$ from $\mathbf{V(\delta_{th})}$, to achieve the Gap statistic $\mathbf{Gap_n(\delta_{th})}$.

### 2.6.4. Gap statistic performance analysis

For each examined MSR value, the Gap statistic method requires applying our biclustering algorithm m+1 times, where m is the number of time we apply it on the reference distribution to achieve an estimate to $E_n^* V(\delta_{th})$, and the additional run is for the examined dataset $V(\delta_{th})$.

Consider a matrix drawn from the reference distribution. Applying the first stage of our suggested method (FLOC) will result with biclusters' size significantly smaller than in the examined dataset, as no biclusters are embedded in it. Moreover, as we increase the MSR value, biclusters discovered in the examined dataset are much more likely to resemble to the ones discovered in lower MSR values. Proceeding to the second stage of our algorithm we notice that the biclusters on which the CASS is applied are much smaller in size and less overlapped, comparing to the examined dataset.

Unlike the FLOC algorithm, the CASS performs an exhaustive (yet optimized) combinatorial search. Applying it on biclusters discovered from the reference distribution shall therefore result with a longer runtime, as its bidders (biclusters) place much lower and closer bids (biclusters' size) and are less likely to be nested in each other. This phenomenon has bigger impact in higher MSR thresholds, as the number of bidders grows.

Empirically, applying our algorithm on a 1000X2000 matrix, drawn from reference distribution typical for a microarray, took approximately 3 hours for an MSR threshold of $\delta_A/2$ using a dual core 64-bit personal computer, where $\delta_A$ is the MSR of the entire reference matrix. Applying the algorithm with $\delta_A/4$ took approximately a single hour, while $\delta_A/8$ took approximately 25 minutes.

We see that for typical microarray dimensions the suggested Gap statistic analysis runtime tends to increase quite dramatically with the examined MSR value. However, in most cases the maximal Gap can already be detected during the process, as we continuously increase the MSR values we examine, so that the analysis for higher MSR vales may already be redundant.

To make the method more scalable for different applications, an analysis of the typical behavior of $E_n^* V(\delta_{th})$, applied on matrices drawn from different reference distributions, can be considered as future work.

## 3. EXPERIMENTS

In order to validate our method we conduct a series of experiments on synthetic and real-world data. In the synthetic experiments we generate uniformly distributed matrices, in which we embed exclusive row submatrices. Our goal is to discover these submatrices using our proposed algorithm and compare it to biclusters discovered by the designated FLOC algorithm. In the real-world data experiment we focus on a Leukemia data set provided by Armstrong et al. [18]. In their paper, Armstrong et al. showed that differences in gene expression data allow us to classify three types of Leukemia (Mixed Linkage Leukemia (MLL), Acute Lymphoblastic Leukemia (ALL) and Acute Myelogenous Leukemia (AML)). Their data set consists of expression data from Affymetrix chips for 12600 genes collected from 72 Leukemia patients, of whom 28 were diagnosed with ALL, 20 were MLL patients, and 24 were AML patients. Our goal in this experiment is therefore to correctly cluster these three types of Leukemia patients and find the genes that correspond to each them. We compare our results with the results obtained by Sheng at el. [5], who applied a Gibbs Sampling biclustering method on the same data set. Notice that in this experiment a single patient is assumed to carry only one type of Leukemia but a gene can influence several types of them. This justifies our interest in exclusive row biclustering for this problem.

### 3.1 Synthetic Data Experimental Results

In the first synthetic data experiment our goal is to extract ten 10X10 exclusive row perfect (zero MSR) biclusters from a 100X50 matrix, where entries that are not part of any bicluster are drawn from a uniform distribution. We first apply our algorithm on this matrix using different values of MSR threshold $\delta > 0$, and explore the outcome of each stage of our suggested method. We then apply the Gap statistic method to indentify the optimal MSR threshold. For convenience, we use the notation $\delta_A$ to represent the MSR of the entire matrix A, and define the MSR threshold to be a fraction of it. Notice that by setting $\delta$ to be as small as possible we are supposed to extract all the perfect biclusters already in the first stage. However, since FLOC suffers from inherent sub optimality it cannot guarantee to discover all those biclusters.

We start with a graphic representation of our biclustering experiment's results. Figure 7 presents the original 100X50 matrix, with the ten perfect biclusters represented by the shaded areas. The quantitative values are hidden for presentation convenience reasons. Notice that a bicluster is not necessarily constructed by consecutive columns, as in bicluster no. 7. The dashed bordered areas on the left hand side illustrate different $\delta$-biclusters discovered in the first stage by the FLOC algorithm, for an MSR threshold value of $\delta_A/20$.

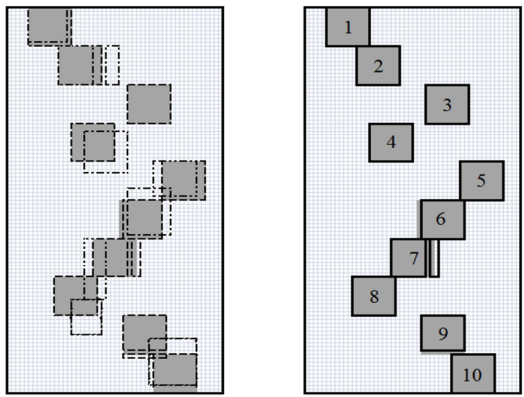

**Figure 7**. *Biclusters discovered (on the left) and chosen (on the right) in the first synthetic data experiment*

Notice that the FLOC algorithm discovers overlapping δ-biclusters, not always at the same position as the original perfect biclusters.

In the second stage we apply the CASS algorithm on the δ-biclusters discovered in the first stage. The chosen biclusters are presented with the bold bordered biclusters on the right hand side of figure 7. We can see that CASS finds the optimal δ-biclusters such that the sum of volumes is maximized with no overlap in rows. We can also see that even though the first stage provides some "false" δ-biclusters, in terms of how close they are to the embedded ones, the second stage "corrects" it by discarding those biclusters, and eventually selecting those which achieve higher global volume and are closer to the embedded biclusters. We examine the performance of our algorithm on the same synthetic data with different values of MSR threshold, as summarized in Table 1.

**Table 1.** Synthetic data results for different MSR threshold values

| MSR threshold | Number of discovered biclusters, for only the specified threshold value | Pct. of correctly clustered rows, for only the specified threshold value | Pct. of correctly clustered rows applying the complete method | Pct. of correctly clustered columns applying the complete method | Pct. of correctly clustered rows, applying only FLOC (K=10) | Pct. of correctly clustered columns, applying FLOC (K=10) |
|---|---|---|---|---|---|---|
| $\delta_A/50$ | 6 | 34 | 34 | 30 | 30 | 25 |
| $\delta_A/30$ | 8 | 80 | 80 | 78 | 50 | 55 |
| $\delta_A/20$ | 10 | 99 | 99 | 99 | 40 | 40 |
| $\delta_A/18$ | 9 | 90 | 100 | 99 | 30 | 35 |
| $\delta_A/16$ | 9 | 90 | 100 | 100 | 30 | 35 |
| $\delta_A/14$ | 8 | 80 | 100 | 100 | 30 | 35 |
| $\delta_A/12$ | 8 | 80 | 100 | 100 | 30 | 40 |
| $\delta_A/10$ | 7 | 70 | 100 | 100 | 30 | 30 |
| $\delta_A/8$ | 7 | 68 | 100 | 100 | 20 | 25 |
| $\delta_A/6$ | 7 | 58 | 100 | 100 | 20 | 20 |
| $\delta_A/4$ | 5 | NA | NA | NA | 10 | 5 |

The first two columns refer to a partial version of our algorithm, which does not include the step 1.b. This is to emphasize on the effect of additionally searching for δ-biclusters with δ values smaller than the requested threshold, as described in section 2.4. The percentage of correctly clustered rows, applying the complete algorithm, can be found on the third column, and the results achieved by the FLOC algorithm with K=10 are described in the column on the right.

We see that for smaller MSR values, the limited version of our algorithm discovers only a smaller portion of correct biclusters. This is due to the FLOC biclustering extraction limitations which are discussed above. On the other hand, when setting a higher value of MSR threshold we again face challenging results, this time since the FLOC algorithm finds biclusters which are significantly bigger than the perfect biclusters. However, these bigger biclusters tend to overlap in their rows and are therefore discarded in the second stage. We can also see that our full method, which allows searching for biclusters with smaller MSR threshold values than the one specified as its input, is able to correctly cluster all rows. By that, it shows to be quite robust to the selection of the MSR threshold, as we notice that by slowly increasing it we can identify saturation in the results it obtains. Comparing to the FLOC algorithm we see that our method demonstrates significantly better performance as it considers the exclusive row nature of the data set.

We now apply the Gap statistic method, to estimate the optimal MSR threshold. We examine 50 MSR threshold values, in the range of $[0,\delta_A/2]$. Figure 8 presents the results we achieve. The synthetic data line refers to the volume we achieve for each of the examined MSR threshold values, while the reference data line describes the discovered volume applying our algorithm on a dataset drawn from a reference distribution. The Gap statistic bars show the corresponding Gap statistic values, $\text{Gap}_n(\delta_{th})$. We first notice the monotonically non-decreasing behavior of both the synthetic and reference datasets. Moreover, we can see the rapid growth in the discovered volume in the synthetic data, compared to the reference data. This growth tends to flatten as we discover an overall volume of approximately 1000, which is exactly the volume we embedded. Looking at the Gap statistic bar, we notice that we achieve the maximal Gap exactly at an MSR index of 5, corresponding to an MSR value of $\frac{5}{50} \cdot \frac{\delta_A}{2} = \frac{\delta_A}{20}$. This value is the threshold for which we first discover all ten biclusters, as we can see in Table 1.

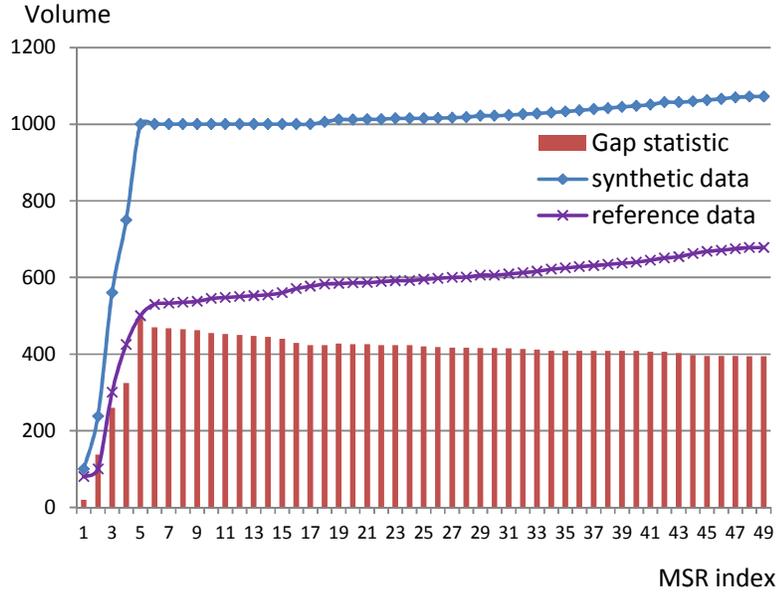

**Figure 8.** *Gap statistic for first synthetic data experiment*

We next demonstrate our method on a more challenging data set. In this experiment we embed five non perfect 200X100 exclusive row biclusters in a 1000X1000 matrix. Each bicluster suffers from a different level of additive noise, such that first bicluster is the least noisy one and the fifth had the highest level of noise. We again apply our method with different values of MSR threshold and apply the Gap statistic method. We compare our results to the FLOC algorithm. Table 2 summarizes our results.

**Table 2.** Second synthetic data experiment results for different MSR threshold values

| MSR Threshold | No. of extracted biclusters | Number of correctly clustered rows (out of 200 rows in each bicluster) | | | | | Overall pct. of correctly clustered rows | Overall pct. of correctly clustered columns | Overall pct. of correctly clustered rows, applying FLOC (K=5) | Overall pct. of correctly clustered columns, applying FLOC (K=5) |
|---|---|---|---|---|---|---|---|---|---|---|
| | | First bicluster | Second bicluster | Third bicluster | Forth bicluster | Fifth bicluster | | | | |
| $\delta_A/30$ | 0 | 0 | 0 | 0 | 0 | 0 | 0 | 0 | 0 | 0 |
| $\delta_A/20$ | 1 | 193 | 0 | 0 | 0 | 0 | 19.3 | 18.5 | 19.3 | 20.1 |
| $\delta_A/14$ | 2 | 197 | 252/400 | | 0 | 0 | 19.7 | 18.7 | 22.4 | 23 |
| $\delta_A/12$ | 4 | 197 | 185 | 192 | 134/400 | | 57.4 | 52.4 | 30.2 | 30.4 |
| $\delta_A/10$ | 4 | 197 | 190 | 192 | 138 | 120 | 83.7 | 77.9 | 31.5 | 32 |
| $\delta_A/8$ | 5 | 197 | 191 | 193 | 158 | 134 | 87.3 | 85.0 | 33.2 | 34.6 |
| $\delta_A/6$ | 5 | 197 | 191 | 194 | 170 | 140 | 89.2 | 90.5 | 31.0 | 35.5 |

| | | | | | | | | | |
|---|---|---|---|---|---|---|---|---|---|
| $\delta_A/4$ | 5 | 197 | 192 | 194 | 185 | 188 | 95.3 | 95.7 | 20.2 | 23.3 |
| $\delta_A/3$ | 5 | 197 | 192 | 194 | 189 | 190 | 96.2 | 97.2 | 19 | 25.7 |

We can see that by increasing the MSR threshold value we are able to correctly cluster rows from noisier biclusters and achieve greater discovered volume. We also note that the results we get are significantly better than when applying the FLOC algorithm, even with the given desired number of biclusters as an input.

Applying the Gap statistic we get the following results, described in Figure 9.

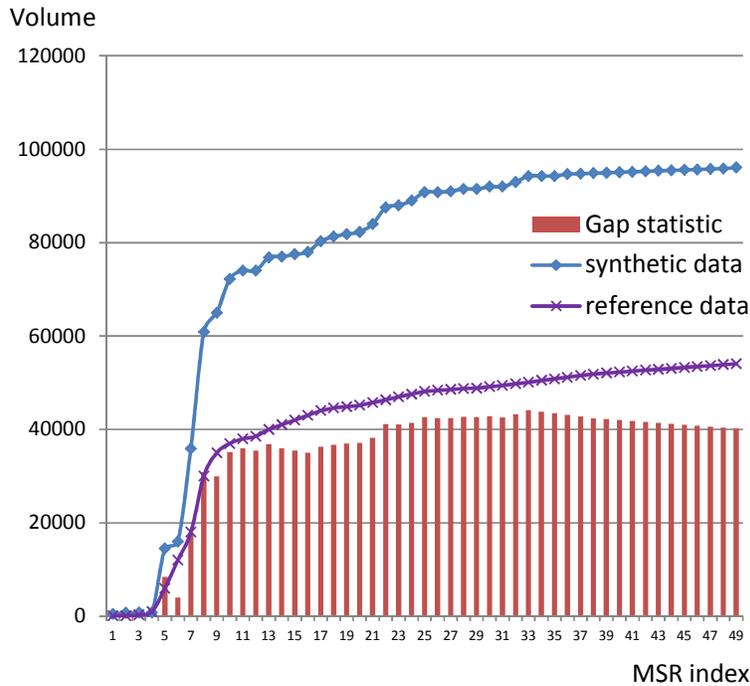

**Figure 9.** *Gap statistic for second synthetic data experiment*

We again see the monotonically non-decreasing behavior, which is less steep than in the previous example, as different biclusters have different MSR threshold values. We also notice the local maximas in the Gap statistic at different elbows in the graph, while the global maxima is achieved at an MSR index of 33, corresponding to a threshold value of $\frac{\delta_A}{3}$. This value is again the MSR value for which we discovered the five embedded biclusters, as can be verified in Table 2.

## 3.2 Leukemia Data Experimental Results

We now turn to examine our suggested method on real-world data, in the form a Leukemia patients data set. This data set contains 12600 gene expression values, collected from 72 Leukemia patients of whom 28 were diagnosed with ALL, 20 were MLL patients, and 24 were AML patients. Our goal is to correctly cluster patients according to their Leukemia type and by that to identify similar expression behavior over a subset of genes.

As a preprocessing step, we follow Sheng at al. [5] who also analyzed the same data set and determined an upper and lower threshold bound on the gene expression values. We define a lower threshold of 100 and an upper threshold of 1600 to overcome noise and saturation measurement effects, respectively. We also examine the variation of each gene along all patients to identify only those genes that significantly vary and may be more valuable for differentiating between patients. We choose to select only the first 15 percent of genes with the highest standard deviation (as in Sheng at el.) which reduces our data set to contain 1887 genes for 72 Leukemia patients.

We first examine our method on various values of MSR threshold in the same manner as we did in the previous section. The obtained results are summarized in Table 3.

**Table 3.** Real-world data results for different MSR threshold values

| MSR Threshold | Number of discovered biclusters | Number of correctly clustered patients | | |
|---|---|---|---|---|
| | | ALL Leukemia | MLL Leukemia | AML Leukemia |
| $\delta_A/20$ | 0 | 0 | 0 | 0 |
| $\delta_A/15$ | 1 | 14/28 | 0 | 0 |
| $\delta_A/10$ | 2 | 24/28 | 28/44 | |
| $\delta_A/8$ | 2 | 24/28 | 32/44 | |
| $\delta_A/7$ | 2 | 25/28 | 33/44 | |
| $\delta_A/6$ | 3 | 25/28 | 17/20 | 19/24 |
| $\delta_A/5$ | 3 | 26/28 | 17/20 | 21/24 |
| $\delta_A/4$ | 2 | 26/28 | 39/44 | |
| $\delta_A/3$ | 2 | NA | NA | NA |
| $\delta_A/2$ | 2 | NA | NA | NA |

We see that for lower MSR threshold values we hardly discover any biclusters, which is due to the fact that the biclusters embedded in the data set are not perfect in real world data. We can also see that by increasing the value of the MSR threshold we first discover the ALL Leukemia patients as a separate cluster, with a high accuracy rate, while the other two Leukemia type patients are still clustered together. As we continue to increase the MSR threshold range we are able to distinguish between the other two Leukemia patient groups and attain three separate Leukemia type biclusters with 89% of the patients correctly clustered, and no false clustering at all.

As we further increase the MSR threshold we see larger yet less accurate biclusters which do not correctly group the Leukemia patients. This is explained by a too large MSR threshold value relatively to the data set.

Figure 10 shows the results we achieve, applying the Gap statistic method to find the optimal MSR threshold.

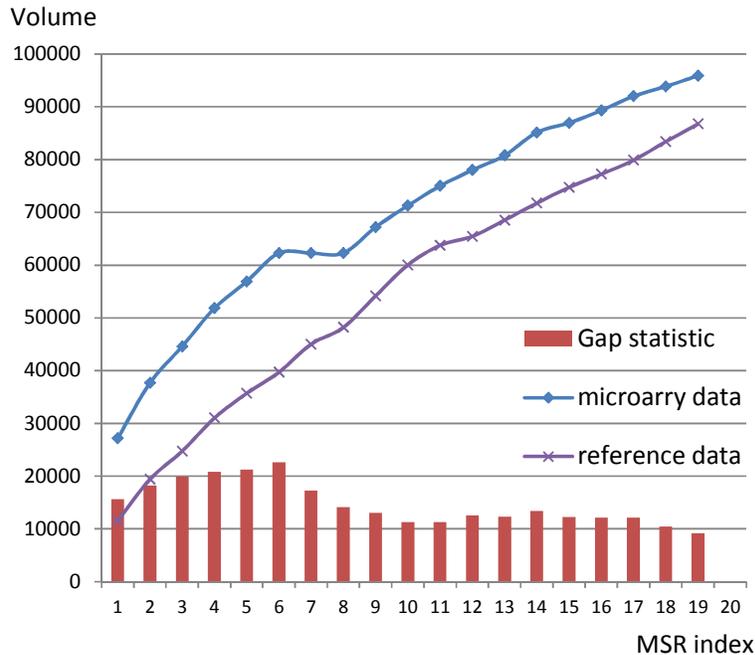

**Figure 10.** *Gap statistic for real-world data experiment*

As in the synthetic data experiment, we see that the maximal Gap is achieved at the location of the elbow, which correspond to an MSR value of $\frac{6}{20} \cdot \frac{\delta_A}{2} = \frac{3\delta_A}{20}$. Comparing this result against table 3, we see that this value indeed indicates the first appearance of three separate biclusters, corresponding to the three Leukemia type patients.

Sheng at el. analyzed the same data set using a Gibbs Sampling approach. They performed sequential biclustering and forced exclusive row structure by masking the discovered rows in each step to avoid row overlap. Applying their algorithm on the Leukemia patients data set, and given the number of desired biclusters, they were able to correctly cluster 76% of the ALL patients, 85.7% of the AML patients and 82.3% of the MLL patients they analyzed. This summarizes to a total of 81% correctly clustered patients. Our two solutions, the one selected by the Gap statistic and the one corresponding to an MSR threshold of $\delta_A/5$, both surpass this accuracy, achieving total accuracy rates of 84.7% and 88.9% respectively.

# 4. SUMMARY

In this work we introduced a novel method to extract maximal volume exclusive row δ-biclusters through a combination of existing MSR biclustering algorithms and a combinatorial auction WDP approach. We showed that our optimization problem can be relaxed and solved in two sequential stages, such that the first stage generates a set of probably overlapping δ-biclusters which are then used as a basis for a volume optimization problem with a no-overlap-in-rows constraint in the second stage. We also introduced a Gap statistic driven approach, to estimate the most suitable MSR threshold value, to set our method's parameter. Through synthetic data simulations we demonstrated the performance of our algorithm and showed its ability to correctly detect the most accurate MSR threshold such that the embedded biclusters are discovered. We also demonstrated our approach on real-world Leukemia gene expression data set. We showed that our algorithm is able to discover the three separate types of Leukemia patients with high accuracy, without prior knowledge of the genes' expression pattern, or the biclusters' expected MSR.

Formulating the exclusive row biclustering problem as a two stage problem, in which the first stage generates candidates for a constrained combinatorial auction WDP optimization, is quite modular to the choice of a biclustering method. In other words, one can choose any type of overlapping biclustering method and apply it in the first stage in our suggested approach. By doing so, every overlapping biclustering method can be applied as exclusive row/column method, while maintaining its other advantages. This ability, together with the theoretical and practical performance we demonstrate, summarizes the main contribution of our work.

# 5. PROOF

Proof of theorem (1): The MSR of a matrix A(I,J) is defined as:

$$H(I, J) = \frac{1}{|I||J|} \sum_{i \in I, j \in J} r(a_{ij})^2 = \frac{1}{|I||J|} \sum_{i \in I, j \in J} (a_{ij} - a_{Ij} - a_{iJ} + a_{IJ})^2$$

Where:

$$a_{iJ} = \frac{1}{|J|} \sum_{j \in J} a_{ij} \ , \ a_{Ij} = \frac{1}{|I|} \sum_{i \in I} a_{ij} \ , \ a_{IJ} = \frac{1}{|I||J|} \sum_{j \in J, i \in I} a_{ij}.$$

Define a column vector of ones, $\mathbf{1} = [1\ 1\ ....1]^T$ and a column vector of zeros, with the $i^{th}$ entry equals to one, $e_i = [0..010..0]^T$. Therefore:

$$a_{ij} = e_i^T A e_j \ , \ a_{iJ} = \frac{1}{|J|} e_i^T A \mathbf{1} \ , \ a_{Ij} = \frac{1}{|I|} \mathbf{1}^T A e_j \ , \ a_{IJ} = \frac{1}{|I||J|} \mathbf{1}^T A \mathbf{1}.$$

The MSR expectation is therefore:

$$E\left\{\frac{1}{|I||J|} \sum_{i \in I, j \in J} r(a_{ij})^2\right\} = \frac{1}{|I||J|} \sum_{i \in I, j \in J} E\{a_{ij} - a_{Ij} - a_{iJ} + a_{IJ}\}^2 =$$

$$\frac{1}{|I||J|} \sum_{i \in I, j \in J} E\left\{e_i^T A e_j - \frac{1}{|J|} e_i^T A \mathbf{1} - \frac{1}{|I|} \mathbf{1}^T A e_j + \frac{1}{|I||J|} \mathbf{1}^T A \mathbf{1}\right\}^2 =$$

$$\frac{1}{|I||J|} \sum_{i \in I, j \in J} E\left\{\left(e_i^T - \frac{1}{|I|}\mathbf{1}^T\right) A e_j + \left(\frac{1}{|I||J|} \mathbf{1}^T - \frac{1}{|J|} e_i^T\right) A \mathbf{1}\right\}^2$$

Define:

$$\beta_i = e_i^T - \frac{1}{|I|}\mathbf{1}^T \ , \ \gamma_i = \frac{1}{|I||J|} \mathbf{1}^T - \frac{1}{|J|} e_i^T$$

$$E\left\{\frac{1}{|I||J|} \sum_{i \in I, j \in J} r(a_{ij})^2\right\} = \frac{1}{|I||J|} \sum_{i \in I, j \in J} E\{\beta_i A e_j + \gamma_i A \mathbf{1}\}^2 =$$

$$\frac{1}{|I||J|} \sum_{i \in I, j \in J} E\left\{\beta_i (A e_j)(A e_j)^T \beta_i^T + 2\beta_i A e_j \mathbf{1}^T A^T \gamma_i^T + \gamma_i (A\mathbf{1})(A\mathbf{1})^T \gamma_i^T\right\} =$$

$$\frac{1}{|I||J|} \sum_{i \in I, j \in J} \beta_i E(A e_j)(A e_j)^T \beta_i^T + 2\beta_i E\{A e_j \mathbf{1}^T A^T\} \gamma_i^T + \gamma_i E(A\mathbf{1})(A\mathbf{1})^T \gamma_i^T$$

It is easy to show that entries of the matrices $\mathrm{E}(Ae_j)(Ae_j)^T$, $\mathrm{E}\{Ae_j \mathbf{1}^T A^T\}$, $\mathrm{E}(A\mathbf{1})(A\mathbf{1})^T$ are in fact linear combinations of $\mathrm{E}\, a_{ij} a_{mn}$  $\forall i, m \in I, j, n \in J$, which means that the entire expression, $\mathrm{E}\left\{\frac{1}{|I||J|} \sum_{i \in I, j \in J} r(a_{ij})^2\right\}$, is only dependent in second order statistics of the matrix A∎